\documentclass[10pt,twocolumn,letterpaper]{article}
\usepackage{booktabs}
\usepackage{tabularx}
\usepackage{rotating} 

\usepackage{cvpr}              

\usepackage[dvipsnames]{xcolor}

\usepackage{graphicx}

\definecolor{cvprblue}{rgb}{0.21,0.49,0.74}
\usepackage[pagebackref,breaklinks,colorlinks,citecolor=cvprblue]{hyperref}

\def\paperID{} 
\def\confName{CVPR}
\def\confYear{2024}

\title{Efficient Image Restoration through Low-Rank Adaptation and Stable Diffusion XL}

\author{Zhao Haiyang\\
School of Data Science\\
City University of Hong Kong\\
{\tt\small haiyazhao7-c@my.cityu.edu.hk}
}

\begin{document}
\maketitle

\begin{abstract}
In this study, we propose an enhanced image restoration model, SUPIR, based on the integration of two low-rank adaptive (LoRA) modules with the Stable Diffusion XL (SDXL) framework. Our method leverages the advantages of LoRA to fine-tune SDXL models, thereby significantly improving image restoration quality and efficiency. We collect 2600 high-quality real-world images, each with detailed descriptive text, for training the model. The proposed method is evaluated on standard benchmarks and achieves excellent performance, demonstrated by higher peak signal-to-noise ratio (PSNR), lower learned perceptual image patch similarity (LPIPS), and higher structural similarity index measurement (SSIM) scores. These results underscore the effectiveness of combining LoRA with SDXL for advanced image restoration tasks, highlighting the potential of our approach in generating high-fidelity restored images. 

\end{abstract}    
\section{Introduction}
\label{sec:intro}

With the advancement of image restoration technology, it has become feasible to construct models capable of generating ultra-high-quality images while retaining the original semantic information as much as possible. Some of these approaches have proven to be highly effective, such as generative priors and increasing the model scale. Among these, model expansion has been demonstrated to be a significant and efficient technique. For instance, notable advancements have been achieved in tasks like Vision Transformers (ViT) \cite{dosovitskiy2020image} and DALL-E \cite{ramesh2021zero} through the expansion of the model scale. This encourages us to further pursue and develop large-scale intelligent IR models capable of generating ultra-high-quality images. 

The SUPIR \cite{yu2024SUPIR} model has demonstrated extraordinary performance in image restoration, using a novel method of improving image restoration ability through text prompts. The author collected 20 million high-quality, high-definition images containing descriptive text annotations for training SUPIR. SUPIR considers Stable Diffusion XL (SDXL) \cite{podell2023sdxl} as a powerful computational prior, containing 2.6 billion parameters. SDXL utilizes an expanded UNet \cite{ronneberger2015U-Net} backbone network and introduces an image-to-image refinement model for post-processing, allowing it to produce images of superior quality and resolution \cite{dosovitskiy2020image}. After exploring the performance of SUPIR, we want to improve its performance in terms of image details and textures. In this work, we introduce two trained LoRA \cite{hu2021Lora} applied to SDXL to fine-tune model parameters and improve the model's face and landscape restoration performance. To verify the effectiveness of this method, we conducted comprehensive experiments on real-world images and achieved better results on indicators PSNR, SSIM, and LPIPS\cite{zhang2018psnrssimlpips}. The main contribution of this work is to shorten the time for image generation and improve the quality of generated images.

\medskip

\noindent

\section{Related Work}
\label{sec:formatting}

\subsection{Image Restoration}

The purpose of image restoration is to convert degraded images into clean, high-quality images \cite{fan2020neural,jinjin2020pipal,zhang2017learning,zhang2022accurate}. Typical image restoration problems include super-resolution \cite{park2003super,dong2015image,lepcha2023image}, deblurring \cite{ren2023multiscale,zhang2022deep,chen2024hierarchical}, and denoising \cite{kawar2022denoising,zhu2023denoising,xia2023diffir}, but these methods generally have limited generalization ability, making it difficult to handle degraded images in the real world.

Deep learning introduces other architectures and training paradigms, thereby improving image restoration performance. For example, transformer-based models enhance the fidelity and authenticity of restored images \cite{huang2017deep-transformer}. In addition, attention mechanisms \cite{vaswani2017attention} and multi-scale feature extraction techniques are integrated into the restoration framework to better capture fine details of images.

In addition to these methods, diffusion models also receive attention in image restoration tasks. Models such as Denoising Diffusion Implicit Model (DDIM) \cite{song2020DDIM} and Denoising Diffusion Neural Network (DDNM) \cite{wang2022DDNM} iteratively refine images through a series of denoising steps, effectively handling various types of image degradation, and show promising results. These models utilize the power of diffusion processes to gradually improve image quality, making them highly effective in tasks such as denoising, deblurring, and more. However, achieving robust performance in various invisible degradation scenarios still faces challenges. Moreover, to better optimize the image restoration model, researchers propose new loss functions, such as Perceptual Loss, which better capture the details of the image and improve the restoration effect. Future research aims to develop more adaptable and scalable models that can effectively generalize to various types of image degradation encountered in real-world applications. Over time, many models that can handle multiple degradation scenarios emerge, among which two-stage methods show good results, such as DiffBIR \cite{lin2023diffbir} and SUPIR \cite{yu2024SUPIR}.

\subsection{Low Rank Adaptation}
Low Rank Adaptation (LoRA) \cite{hu2021Lora} is an approximate numerical decomposition technique that is particularly useful for large-scale language models. This method involves performing low rank decomposition of feature matrices, which allows for efficient adaptation of pre-trained models. By utilizing low rank decomposition techniques, LoRA can significantly reduce the number of parameters in the feature matrix, leading to decreased memory usage and computational overhead.

The core idea behind LoRA is to insert a low rank adaptation matrix into the model architecture, enabling fast adaptation and efficient fine-tuning without altering the original model's weights. This approach is particularly advantageous in scenarios where computational resources are limited or where rapid model updates are necessary. LoRA achieves this by leveraging the inherent low-rank structure within the feature matrices of large-scale models, which often contain redundant information. By decomposing these matrices into lower-dimensional components, LoRA reduces the complexity of the adaptation process. Furthermore, LoRA's ability to maintain the integrity of the original model's weights ensures that the foundational knowledge embedded within the pre-trained model is preserved. This makes LoRA especially effective in transfer learning scenarios, where the pre-trained model is adapted to new tasks or domains. The low-rank adaptation matrix can be trained with relatively fewer resources compared to re-training the entire model, thus making the process more efficient and cost-effective.

In practical terms, LoRA can be applied to various aspects of model adaptation, including fine-tuning for specific tasks, domain adaptation, and even continual learning. Its flexibility and efficiency make it a valuable tool in the toolkit of machine learning practitioners, particularly when dealing with large and complex models. The reduced computational burden also facilitates experimentation and iteration, allowing researchers and engineers to explore a wider range of configurations and settings. LoRA can offer a blend of efficiency, effectiveness, and practicality. Its application can lead to more responsive and adaptable AI systems, capable of quickly incorporating new information and tasks without the need for extensive computational resources.

\subsection{Stable Diffusion XL}

Diffusion models \cite{ho2020diffusion,rombach2022stable-diffusion,yang2023diffusion} garner significant attention in the field of generative artificial intelligence, delivering state-of-the-art outcomes across various applications, including text-to-image \cite{ko2023text-to-image,saharia2022text-to-image,zhang2023text-to-image} and text-to-video \cite{blattmann2023text-to-video,zhang2023text-to-video,wu2023text-to-video} transformations. These models operate by gradually transforming a simple, structured noise distribution into a complex data distribution through a series of iterative refinement steps. This process enables the generation of high-fidelity images and videos from random noise, making diffusion models a powerful tool for various generative tasks.

Stable diffusion \cite{rombach2022stable-diffusion,croitoru2023diffusion-25,mou2024t2i-25} is particularly influential in text-to-image synthesis, leveraging the Latent Diffusion Model (LDM) to execute diffusion operations within a semantically compressed space. This approach enhances computational efficiency by reducing the dimensionality of the data on which diffusion operations are performed. The core architecture of stable diffusion models centers around U-Net, a convolutional neural network architecture that is well-suited for image restoration tasks. U-Net iteratively denoises random latent codes, supported by text encoders and image decoders, to harmonize text and image generation. The use of text encoders allows the model to understand and incorporate textual descriptions into the generated images, resulting in highly detailed and contextually relevant outputs. However, the computational demands of its multi-step inference process \cite{xiao2017multi-step} become a significant burden, particularly when generating high-resolution images or long video sequences. Each step in the diffusion process requires complex computations, leading to substantial time and resource consumption. This computational overhead poses a challenge for real-time applications and large-scale deployments, where efficiency is crucial. To address these challenges, researchers have introduced various distillation techniques such as Progressive Distillation and Adversarial Distillation \cite{salimans2022progressive-distillation,sauer2023adversarial-distillation}. Progressive Distillation incrementally transfers knowledge from a complex model to a simpler one, maintaining the performance while reducing the number of necessary computation steps. Adversarial Distillation, on the other hand, leverages adversarial training to enhance the quality of the distilled model, ensuring that the simplified model retains the generative capabilities of the original.

SDXL Lightning \cite{lin2024sdxl-lighting} is an enhanced version of the SDXL model that employs progressive adversarial distillation technology, to significantly boost the quality and efficiency of image generation. SDXL Lightning employs an advanced model architecture and adversarial training mechanism to generate high-resolution, detailed images while minimizing computational resources. 
In order to reduce the computational requirements of training diffusion models for high-resolution image synthesis, it has been found that although diffusion models can ignore perceptually unimportant details through undersampling loss terms, they still require expensive function evaluation in pixel space, thereby requiring a large amount of computation time and energy resources. Therefore, a method was introduced to clearly separate the compression learning phase. This method employs an autoencoder model that learns a space that is perceptually equivalent to the image space, but with significantly reduced computational complexity.


\begin{figure*}[ht]
    \centering
    \includegraphics[width=\textwidth]{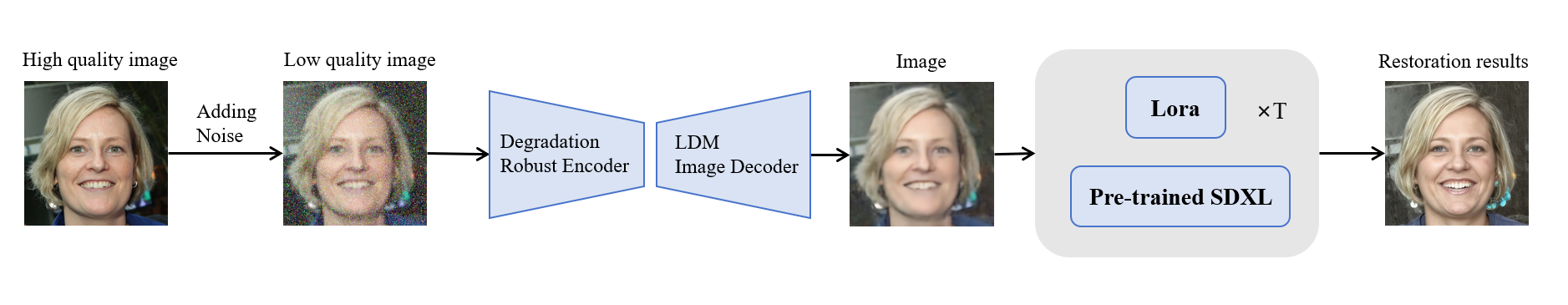}
    \caption{The pipeline of our image restoration model. This figure briefly shows the workflow of the proposed model.}
    \label{fig:pipeline}
\end{figure*}

\section{Method}
\subsection{Background of Stable Diffusion}

The key steps of the stable diffusion model consist of the forward diffusion process and the backward denoising process \cite{ho2020diffusion}. The forward diffusion process progressively adds noise to the data, whereas the backward process utilizes the learned model to remove noise and restore the original data. Specifically, the forward diffusion process can be described as:

\begin{equation}
q(\mathbf{x}_t | \mathbf{x}_{t-1}) = \mathcal{N}(\mathbf{x}_t; \sqrt{\alpha_t} \mathbf{x}_{t-1}, (1 - \alpha_t)\mathbf{I})
\end{equation}

The reverse process is approximated by a parameterized denoising model $p_{\theta}$, which is trained by maximizing logarithmic likelihood estimation:
\begin{equation}
p_{\theta}(\mathbf{x}_{t-1} | \mathbf{x}_t) = \mathcal{N}(\mathbf{x}_{t-1}; \mu_{\theta}(\mathbf{x}_t, t), \sigma^2_{\theta}(\mathbf{x}_t, t)\mathbf{I})
\end{equation}

Among them, $\mu_{\theta}$ and $\sigma_{\theta}$are the learned mean and variance functions, respectively.Its goal is to predict a noise to be added to the input image \( x \) based on the noisy image \( x_t \) at time \( t \). The objective function of LDM \cite{rombach2022stable-diffusion} is

\begin{equation}
\mathcal{L}_{\text{LDM}} = \mathbb{E}_{\mathbf{x}, \epsilon \sim \mathcal{N}(0,1), t} \left[ \left\| \epsilon - \epsilon_\theta(\mathbf{x}_t, t) \right\|_2^2 \right],
\end{equation}

In LDM, we learn in latent space, that is, predict the noise added on $\mathbf{z}_t$, and the corresponding loss function is expressed as follows:

\begin{equation}
\mathcal{L}_{\text{LDM}} = \mathbb{E}_{\mathbf{\xi}_{(x)}, \epsilon \sim \mathcal{N}(0,1), t} \left[ \left| \epsilon - \epsilon_\theta(\mathbf{z}_t, t) \right|_2^2 \right]
\end{equation}

Compared with traditional Generative Adversarial Networks (GANs) \cite{goodfellow2014GANs}, stable diffusion models exhibit enhanced stability and a reduced risk of mode collapse. Moreover, by functioning within the latent space, it substantially enhance the computational efficiency of the generation process.

\subsection{Scaling-UP Image Restoration}

SUPIR combines large-scale pre trained generative models, significantly improving the effectiveness of image restoration. SUPIR adopts a two-stage architecture, with each stage optimized for different tasks. In the first stage, a pre-trained restoration module is employed to remove degraded components from the image, such as blur and noise. In the second stage, SUPIR leverages sdxl for image detail and texture reconstruction.

First, input a low-quality image, and then the low-quality image will be encoded by the fine tuned encoder and mapped to the latent space. This encoder has been specially trained to handle degraded images. The author designed an adapter based on ControlNet \cite{zhang2023Controlnet} that can recognize LQ image content and guide it to recover images based on the provided low-quality input. The adapter adopts a partially trimmed Vision Transformer (ViT) \cite{alexey2020VIT} module and introduces a ZeroSFT \cite{wang2018zeroSFT} module to enhance the guidance effect of LQ images. 

Let the input image be \( \mathbf{I} \). The output of the encoder is \( z_\text{enc} \):
\begin{equation} 
z_\text{enc} = \text{Encoder}(\mathbf{I})
\end{equation}

The encoded feature \( z_\text{enc} \) is processed by the trimmed ControlNet:
\begin{equation}
z_{\text{LQ}} = \text{Conv}(z_\text{enc}) + \text{ZeroConv}(z_\text{enc}, \text{prompt})
\end{equation}
Here, \(\text{ZeroConv}(\cdot)\) represents a convolution operation with zero-padding, ensuring the spatial dimensions of the input are preserved. It is used to integrate additional information (such as prompts) without altering the feature map size. The output of the trimmed ControlNet is then processed by the decoder to generate the final output \( z_\text{dec} \):
\begin{equation}
z_\text{dec} = \text{Decoder}(z_{\text{LQ}})
\end{equation}

The author also introduced the LLaVA \cite{liu2024llava} large language model to clarify the content in low-quality images subjected to robust degradation processing, and output it in the form of text descriptions. Then use these descriptions as prompts to guide the recovery. 
Additionally, the author employs negative prompts to manage the output quality of image generation models, particularly in the absence of Classifier-Free Guidance (CFG) \cite{ho2022CFG}. Specifically, at each step of diffusion, we will make two predictions using positive prompts $\text{pos}$ and negative prompts $\text{neg}$, and take the fusion of these two results as the final output $z_{t-1}$:
\begin{equation}
z_{t-1}^{\text{pos}} = H(z_t, z_{\text{LQ}}, \sigma_t, \text{pos})
\end{equation}

\begin{equation}
z_{t-1}^{\text{neg}} = H(z_t, z_{\text{LQ}}, \sigma_t, \text{neg})
\end{equation}

\begin{equation}
z_{t-1} = z_{t-1}^{\text{pos}} + \lambda_{\text{cfg}} \times (z_{t-1}^{\text{pos}} - z_{t-1}^{\text{neg}})
\end{equation}
where $H(\cdot)$ is our diffusion model with adaptor, $\sigma_t$ is the variance of the noise at time-step $t$, and $\lambda_{\text{cfg}}$ is a hyperparameter. In our framework, $\text{pos}$ can be the image description with positive words of quality, and $\text{neg}$ is the negative words of quality. 
For instance, negative prompts can direct the model to avoid generating blurry, distorted, or low-quality images. Consequently, the SUPIR model is capable of generating high-quality images within the latent space. Subsequently, the produced high-quality images are transformed back into the image space via a fixed decoder. Furthermore, by training on the dataset and leveraging the properties of diffusion models, image restoration is performed selectively based on LLaVA's prompts, effectively addressing a range of restoration requirements. 

In this study, we explore not only the use of SDXL but also several other variants of the Stable Diffusion model. Among these, the SDXL model demonstrates superior performance, while the SDXL Lightning variant also exhibits commendable capabilities. SDXL Lightning is particularly noteworthy for its ability to reduce the number of inference steps to 15 without compromising image quality, thereby enabling the generation of high-quality images within a significantly shorter time frame.

\subsection{Training for Low Rank Adaptation}

In this study, we adopted LoRA technology to adapt the SDXL model and enhance its performance in facial image generation. LoRA effectively fine tunes the pre trained model by introducing a low rank factorization matrix without increasing memory and computational overhead. Specifically, we incorporated a LoRA adaptation layer into the SDXL model and trained two separate LoRa models. One LoRa model was trained on 1300 landscape images, while the second was specifically trained on 300 facial images. All images are preprocessed to a resolution of 512x512 to ensure data consistency and quality. Each image is labeled in detail to ensure learning ability during training. This method has been proven to be efficient and adaptable under resource constrained conditions in multiple studies \cite{biderman2024lora-ok}. During the training process, the LoRA adaptation matrix is initialized to zero and the update matrix is initialized using a random Gaussian distribution to ensure the stability of the model in the early stages of training. We adjust hyperparameters such as learning rate based on specific performance to optimize model performance. Through this low rank adaptation method, we successfully improved the performance of the SUPIR model in facial image generation and validated the effectiveness of LoRA technology in model fine-tuning.

As mentioned earlier, LoRA represents parameter updates through low rank matrix decomposition. Given a weight matrix \( W \), LoRA decomposes it into two low-rank matrices \( A \) and \( B \), such that:

\begin{equation}
\Delta W = A \times B
\end{equation}

where the ranks of \( A \) and \( B \) are much smaller than the original dimensions of \( W \).

During training, LoRA updates only the matrices \( A \) and \( B \), keeping \( W \) unchanged. This method drastically reduces the number of parameters that need to be trained. For a weight matrix \( W \in \mathbb{R}^{d \times k} \), if we choose a rank \( r \ll \min(d, k) \), the number of parameters to be adjusted decreases from \( d \times k \) to \( (d \times r + r \times k) \). The training process can be described as follows:

Initialize low-rank matrices \( A \in \mathbb{R}^{d \times r} \) and \( B \in \mathbb{R}^{r \times k} \). At each iteration, update \( A \) and \( B \) based on the gradient of the loss function:
\begin{equation}
A \leftarrow A - \eta \frac{\partial \mathcal{L}}{\partial A}
\end{equation}
\begin{equation}
B \leftarrow B - \eta \frac{\partial \mathcal{L}}{\partial B}
\end{equation}
where \( \eta \) is the learning rate and \( \mathcal{L} \) is the loss function.

The effective weight matrix during training is given by:
\begin{equation}
W' = W + \Delta W = W + A \times B
\end{equation}
The optimization objective can be expressed as minimizing the loss function with respect to the effective weights:
\begin{equation}
\mathcal{L}(W') = \mathcal{L}(W + A \times B)
\end{equation}
To ensure convergence and stability, regularization terms can be added to the loss function to penalize large updates in \( A \) and \( B \):
\begin{equation}
\mathcal{L}_{\text{reg}} = \mathcal{L}(W') + \lambda (\|A\|_F^2 + \|B\|_F^2)
\end{equation}
where \( \| \cdot \|_F \) denotes the Frobenius norm and \( \lambda \) is a regularization parameter.
By this means, LoRA achieves efficient parameter updates and demonstrates superior performance across various tasks, such as image and text generation, while significantly reducing computational and storage costs.


\begin{figure*}[htbp]
  \centering
  \begin{subfigure}[b]{0.24\textwidth}
    \centering
    \includegraphics[width=\textwidth]{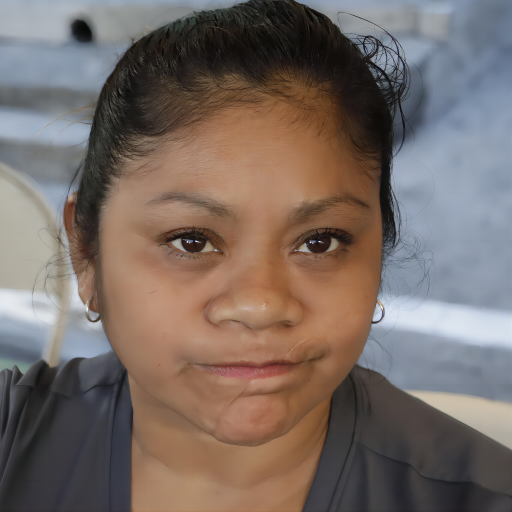}
    \label{fig:11}
  \end{subfigure}
  \begin{subfigure}[b]{0.24\textwidth}
    \centering
    \includegraphics[width=\textwidth]{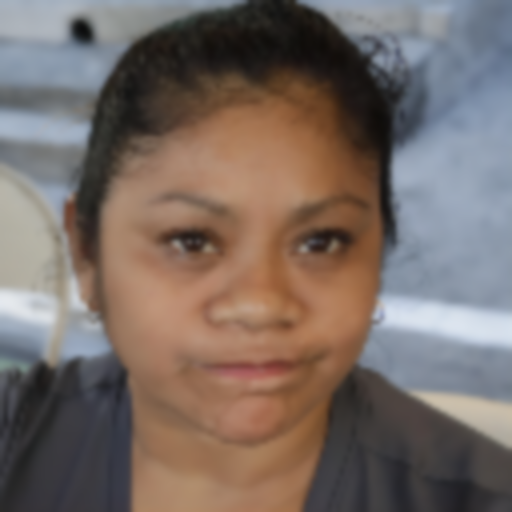}
    \label{fig:12}
  \end{subfigure}
  \begin{subfigure}[b]{0.24\textwidth}
    \centering
    \includegraphics[width=\textwidth]{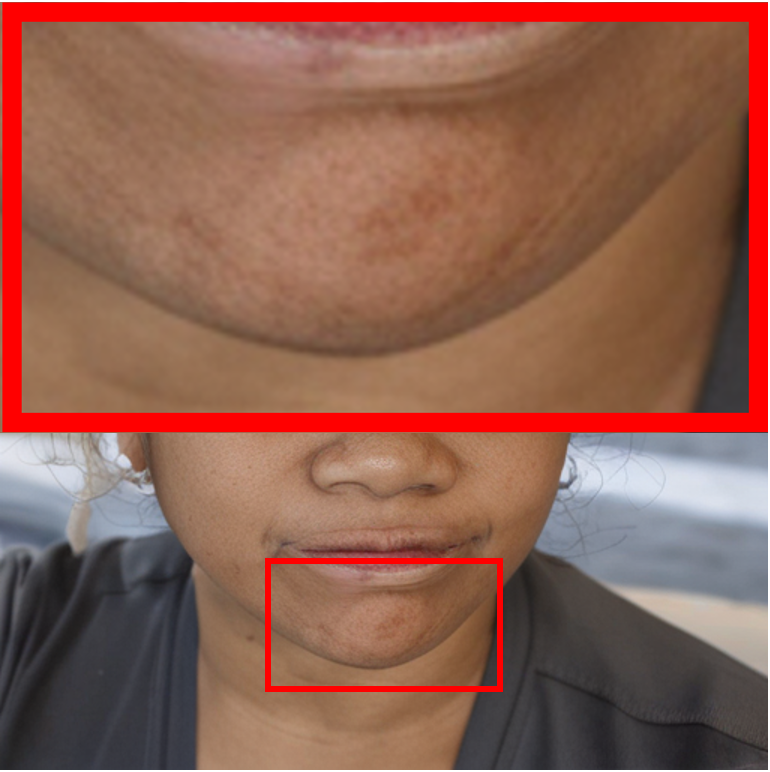}
    \label{fig:13}
  \end{subfigure}
  \begin{subfigure}[b]{0.24\textwidth}
    \centering
    \includegraphics[width=\textwidth]{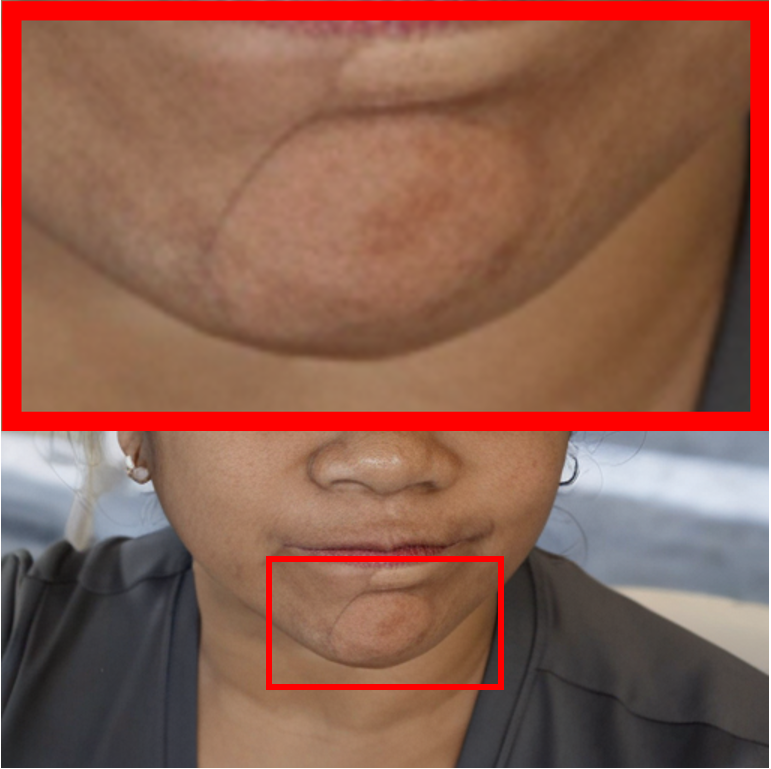}
    \label{fig:14}
  \end{subfigure}
  
  \vspace{-0.2cm}
  \vspace{-0.2cm}
 
  \begin{subfigure}[b]{0.24\textwidth}
    \centering
    \includegraphics[width=\textwidth]{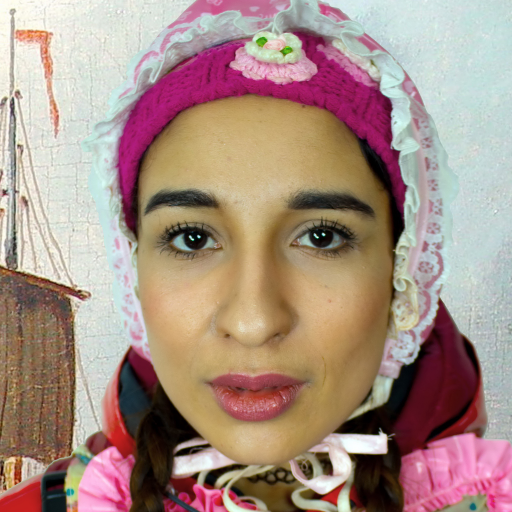}
    \label{fig:21}
  \end{subfigure}
  \begin{subfigure}[b]{0.24\textwidth}
    \centering
    \includegraphics[width=\textwidth]{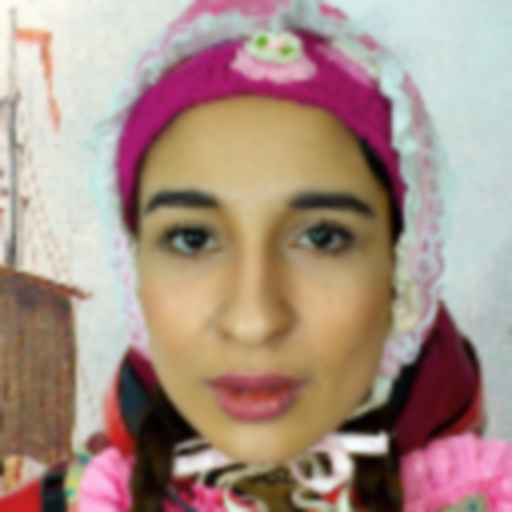}
    \label{fig:22}
  \end{subfigure}
  \begin{subfigure}[b]{0.24\textwidth}
    \centering
    \includegraphics[width=\textwidth]{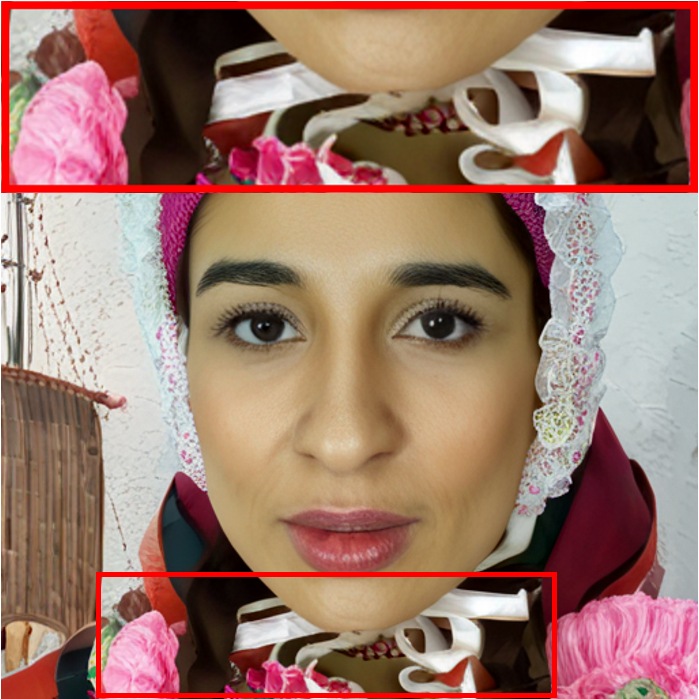}
    \label{fig:23}
  \end{subfigure}
  \begin{subfigure}[b]{0.24\textwidth}
    \centering
    \includegraphics[width=\textwidth]{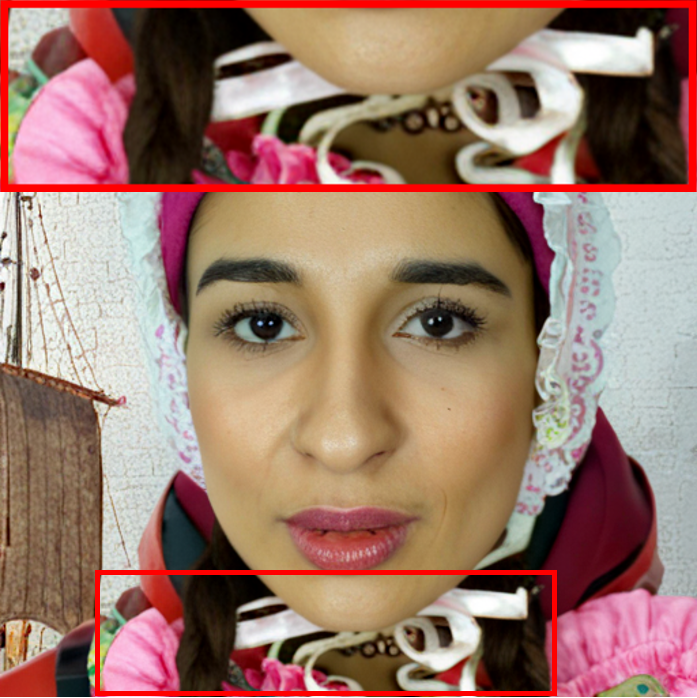}
    \label{fig:24}
  \end{subfigure}
   
  \vspace{-0.2cm}
  \vspace{-0.2cm}
  
  \begin{subfigure}[b]{0.24\textwidth}
    \centering
    \includegraphics[width=\textwidth]{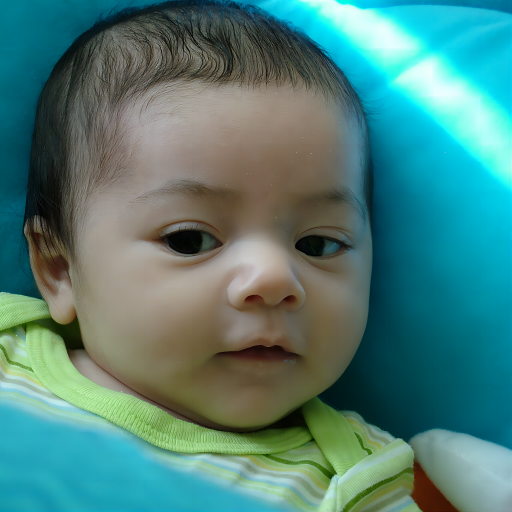}
    \caption{GT}
    \label{fig:31}
  \end{subfigure}
  \begin{subfigure}[b]{0.24\textwidth}
    \centering
    \includegraphics[width=\textwidth]{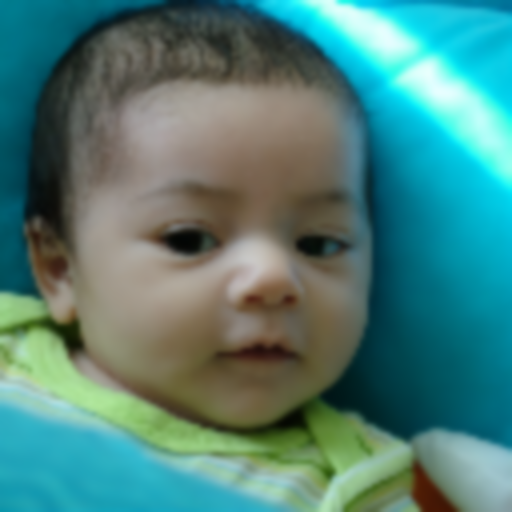}
    \caption{Low-quality image}
    \label{fig:32}
  \end{subfigure}
  \begin{subfigure}[b]{0.24\textwidth}
    \centering
    \includegraphics[width=\textwidth]{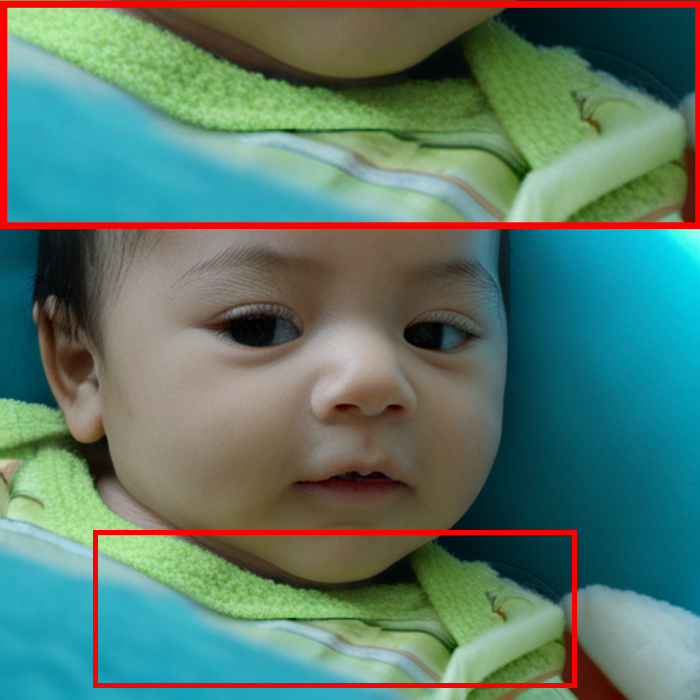}
    \caption{SUPIR}
    \label{fig:33}
  \end{subfigure}
  \begin{subfigure}[b]{0.24\textwidth}
    \centering
    \includegraphics[width=\textwidth]{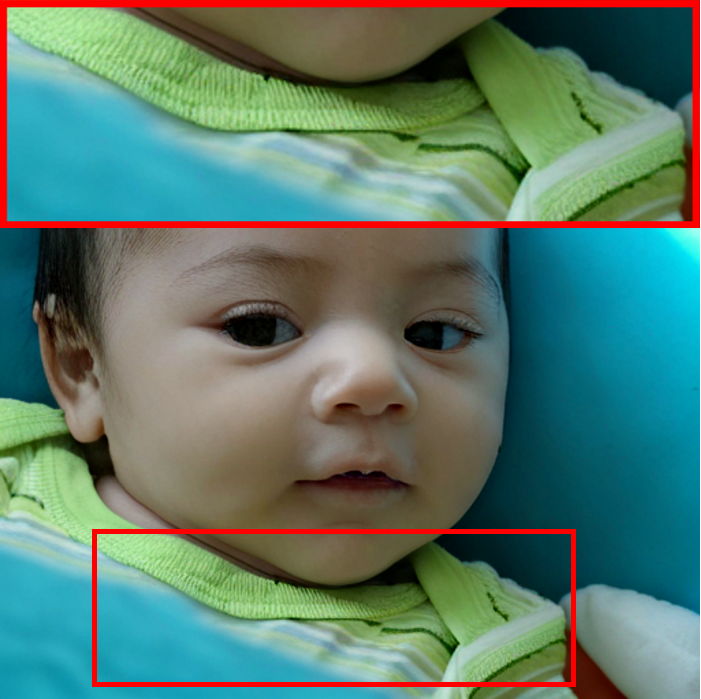}
    \caption{Ours}
    \label{fig:34}
  \end{subfigure}
  
  \caption{Compare with SUPIR. We apply a mixture of Gaussian blur with $\sigma=2$ and 4$\times$ downsampling for super-resolution degradation. Our method has a good restoration effect on facial details, such as scars. For the texture of hair and clothing, our model has a stronger effect than SUPIR.}
  \label{fig:enter-label}
\end{figure*}

\section{Experiments}
\subsection{Model Training and Quantitative Comparisons}

\begin{table}
  \centering
  \scriptsize 
  \setlength{\tabcolsep}{4pt} 
  \renewcommand{\arraystretch}{1.5} 
  \begin{tabular}{@{}lcccc@{}}
    \toprule
    Method & Degradation & PSNR$\uparrow$ & SSIM$\uparrow$ & LPIPS$\downarrow$ \\
    \midrule
    
    \parbox[t]{3cm}{(Blur ($\sigma=3$) + Noise ($\sigma=30$))} & Ours & \textbf{32.19} & \textbf{0.7434} & \textbf{0.0932} \\
    & Lighting-LoRA & 29.37 & 0.5834 & 0.1232 \\
    & HWXL-LoRA & 28.87 & 0.6025 & 0.1183 \\
    & SUPIR & 29.46 & 0.4203 & 0.1402 \\
    & Lighting & 29.63 & 0.5523 & 0.2085 \\
    & HWXL & 29.13 & 0.5856 & 0.1490 \\
    
    \hline
    SR ($\times 4$) & Ours & 29.64 & \textbf{0.6382} & \textbf{0.0916} \\
    & Lighting-LoRA & 29.03 & 0.5795 & 0.1328 \\
    & HWXL-LoRA & 28.78 & 0.6004 & 0.1265 \\
    & SUPIR & \textbf{29.81} & 0.5934 & 0.1432 \\
    & Lighting & 28.57 & 0.5357 & 0.2250 \\
    & HWXL & 28.64 & 0.5774 & 0.1688 \\
    \hline
    Blur ($\sigma=2$) + SR ($\times 4$) & Ours & \textbf{29.38} & \textbf{0.5651} &\textbf{0.1250} \\
    & Lighting-LoRA & 28.11 & 0.5436 & 0.1414 \\
    & HWXL-LoRA & 28.65 & 0.5609 & 0.1293 \\
    & SUPIR & 27.75 & 0.4702 & 0.1306 \\
    & Lighting & 27.33 & 0.4694 & 0.2880 \\
    & HWXL & 28.78 & 0.5495 & 0.1803 \\
    \hline
    \parbox[t]{3cm}{Blur ($\sigma=2$) + SR ($\times 4$) + Noise ($\sigma=1$)} & Ours & 18.48 & \textbf{0.2881} & 0.3505 \\
    & Lighting-LoRA & 17.44 & 0.2590 & 0.4075 \\
    & HWXL-LoRA & 19.70 & 0.2705 & \textbf{0.2907} \\
    & SUPIR & 18.57 & 0.2808 & 0.3225 \\
    & Lighting & 17.39 & 0.1681 & 0.6379 \\
    & HWXL & \textbf{20.94} & 0.2823 & 0.4472 \\
    
    \bottomrule
  \end{tabular}
  \caption{Comparison of data between different methods under different degradation. \textbf{Bold} represents the best performance. ↓ represents the smaller the better, and for the others, the bigger the better.}
  \label{tab:1}
\end{table}

For training, We use the AdamW optimizer \cite{loshchilov2017AdamW-optimizer} with a learning rate of 0.0001. The training process spans 2 days, with a batch size of 256. In our experiments, the integration of two LoRA modules with the SDXL model in the SUPIR framework demonstrated remarkable improvements in image restoration tasks. We used three SDXL models, namely SDXL, SDXL-lighting, HelloWorld-XL, Among them, HelloWorld XL used 20821 images as the training set, which includes different people and actions, as well as many lifelike animals. Moreover, the image quality of close-up portrait output is better than SDXL. And HelloWorld-XL intentionally includes some low-quality images in the training to enhance the model's response to negative prompts, which is also why HelloWorld-XL performs well in handling blur and noise. The proposed method achieved excellent performance across multiple metrics, with Peak Signal-to-Noise Ratio (PSNR) values significantly higher than the baseline models, indicating clearer and more accurate restored images. Additionally, the Learned Perceptual Image Patch Similarity (LPIPS) scores were notably lower, reflecting a better perceptual similarity to the ground truth images. Furthermore, the Structural Similarity Index Measure (SSIM) scores were substantially improved, showcasing enhanced structural fidelity and visual quality of the restored images. These outstanding results affirm the effectiveness of our approach in producing high-quality image restorations, making it a promising solution for advanced image restoration applications.

To generate low-quality images to test the performance of our method, we introduced various degradations ranging from simple to complex. For quantitative comparison, we selected the following indicators: complete reference indicators PSNR, SSIM, and LPIPS \cite{zhang2018psnrssimlpips}. Compared with the original SUPIR method, our method has improved in all parameter indicators. Similarly, in \cref{fig:enter-label}, it can be seen that our model has indeed achieved good results in face restoration, with some progress compared to SUPIR in certain small details and colors

In terms of details in image restoration, our method demonstrates some better features than the original SUPIR model. For example, in \cref{fig:first} we can see that the texture of the goat's wool on the trained Lora image is more in line with the texture of the original image. In the image of the little girl, we can see that low-quality images basically do not show the earrings. In the SUPIR model, the earrings are also restored to hair, while the image generated by the trained Lora will show the earrings. Therefore, it can be demonstrated that our method generates high fidelity textures.


\begin{table}
  \centering
  \scriptsize
  \setlength{\tabcolsep}{2pt} 
  \begin{minipage}{0.46\textwidth} 
    \begin{tabularx}{\textwidth}{@{}l*{5}{>{\centering\arraybackslash}X}@{}}
    \toprule
    Metrics & PASD & Stable-SR& DiffBIR& SUPIR & Ours \\
    \midrule
    PSNR$\uparrow$ & 26.87 & 19.76 & 28.72 & 27.74 & \textbf{29.38} \\
    SSIM$\uparrow$ &  0.4513 & 0.4051 & 0.4663 & 0.4702 & \textbf{0.5651} \\
    LPIPS$\downarrow$ & 0.1828 & 0.2418 & 0.1289 & 0.1306 & \textbf{0.1250} \\
    \bottomrule
  \end{tabularx}
  \caption{Quantitative comparisons on 60 real-world images.}
  \label{tab:2}
  \end{minipage}
\end{table}


\begin{table}
  \centering
  \scriptsize
  \setlength{\tabcolsep}{2pt} 
  \begin{minipage}{0.46\textwidth} 
    \begin{tabularx}{\textwidth}{@{}l*{5}{>{\centering\arraybackslash}X}@{}}
      \toprule
      Metrics & PASD & Stable-SR & DiffBIR & SUPIR & Ours\\
      \midrule
      Computational Time & 28.65 & 1013 & 45.20 & 18.44 & \textbf{11.28}\\
      \bottomrule
    \end{tabularx}
    \caption{Quantitative comparison of time using different methods.}
    \label{tab:3}
  \end{minipage}
\end{table}

\begin{figure*}[htbp]
  \centering
  \begin{subfigure}[b]{0.19\textwidth}
    \centering
    \includegraphics[width=\textwidth]{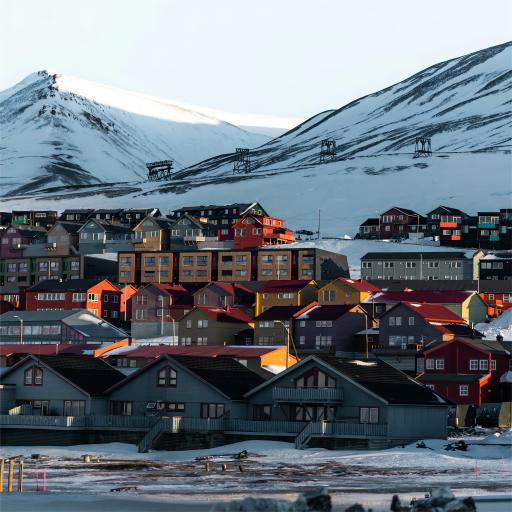}
    \label{fig:41}
  \end{subfigure}
  \begin{subfigure}[b]{0.19\textwidth}
    \centering
    \includegraphics[width=\textwidth]{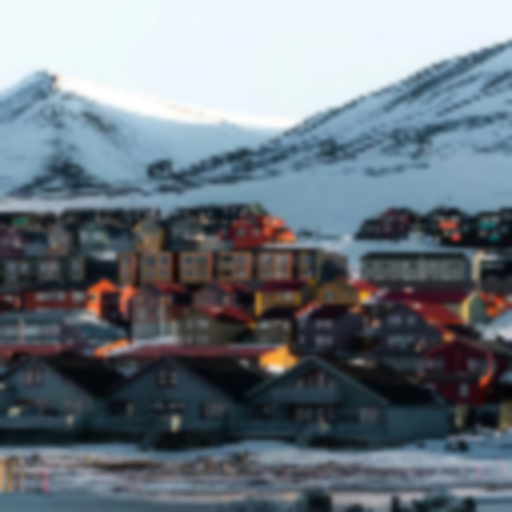}
    \label{fig:42}
  \end{subfigure}
  \begin{subfigure}[b]{0.19\textwidth}
    \centering
    \includegraphics[width=\textwidth]{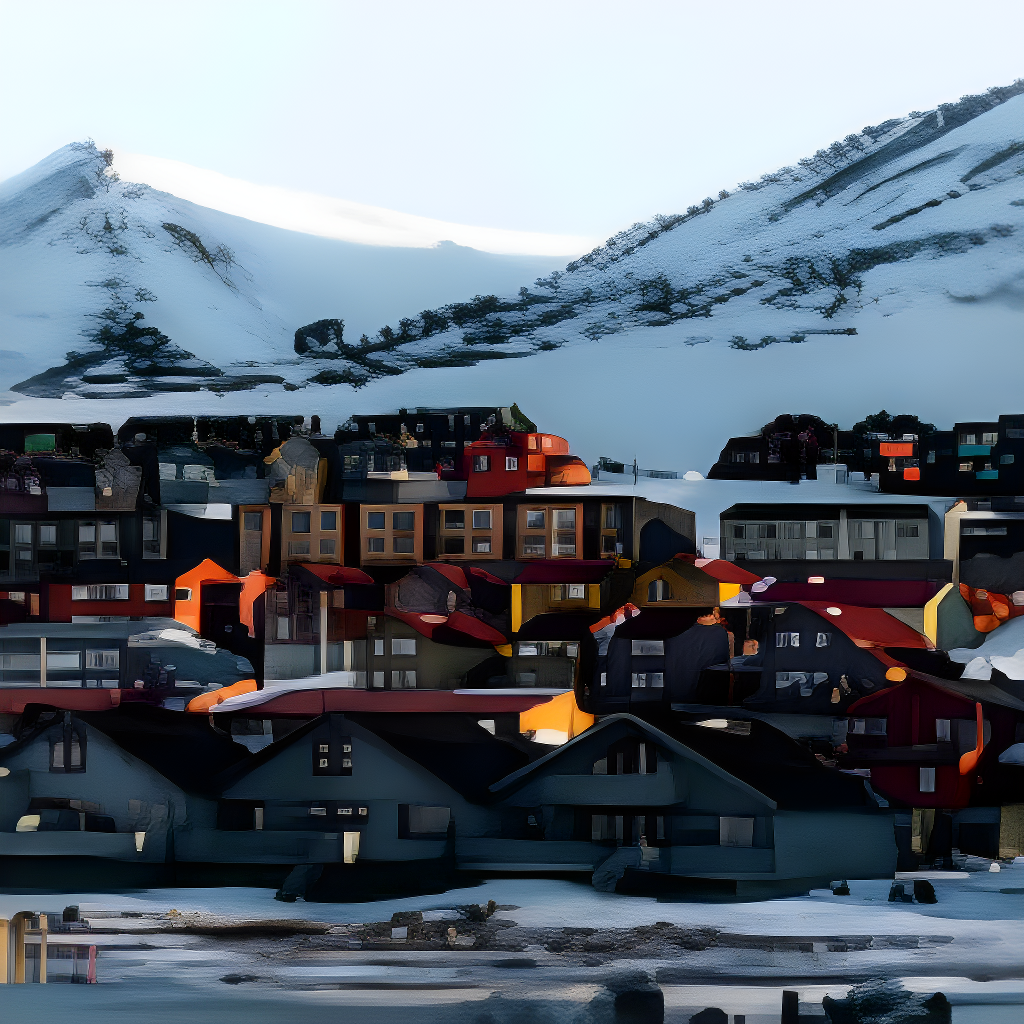}
    \label{fig:43}
  \end{subfigure}
  \begin{subfigure}[b]{0.19\textwidth}
    \centering
    \includegraphics[width=\textwidth]{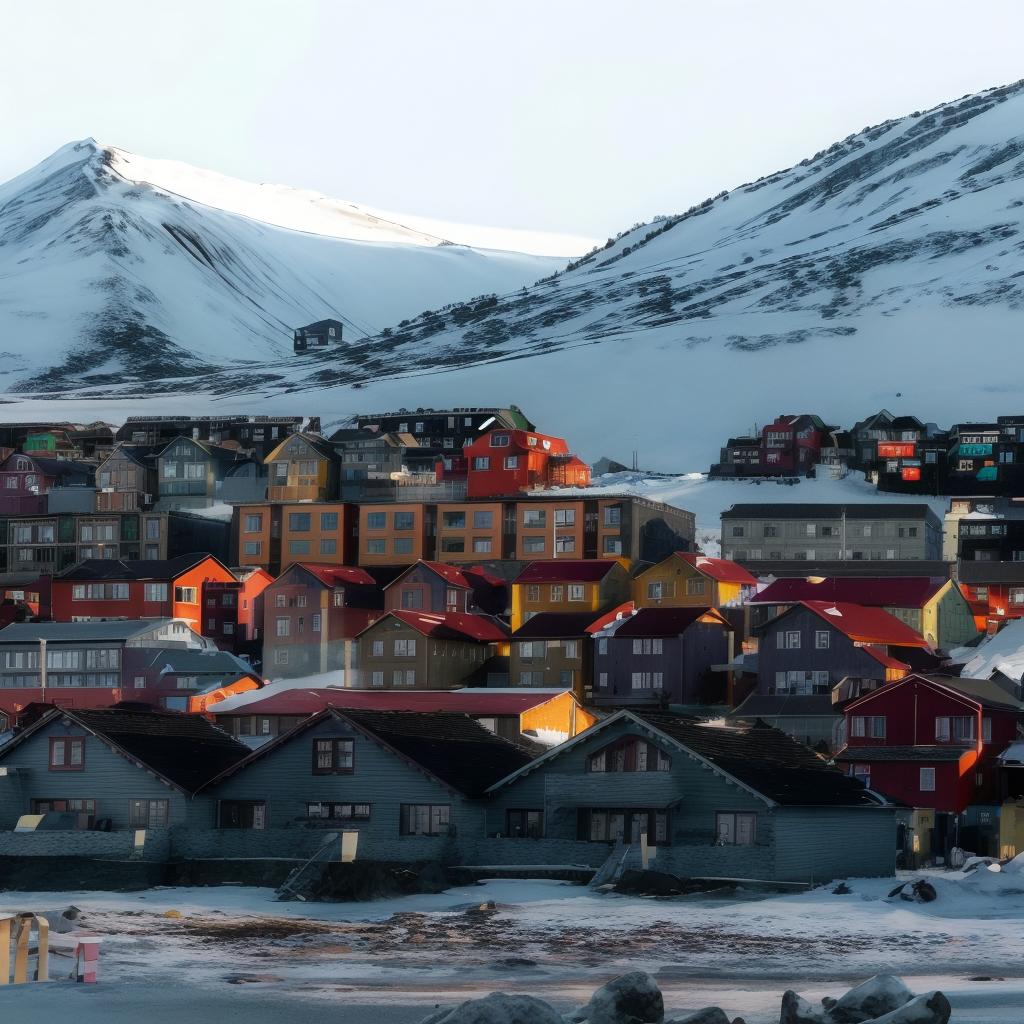}
    \label{fig:44}
  \end{subfigure}
  \begin{subfigure}[b]{0.19\textwidth}
    \centering
    \includegraphics[width=\textwidth]{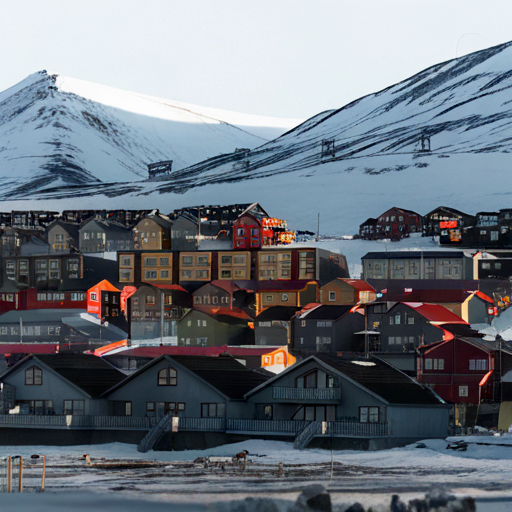}
    \label{fig:45}
  \end{subfigure}

  \vspace{-0.2cm}
  \vspace{-0.2cm}

  \begin{subfigure}[b]{0.19\textwidth}
    \centering
    \includegraphics[width=\textwidth]{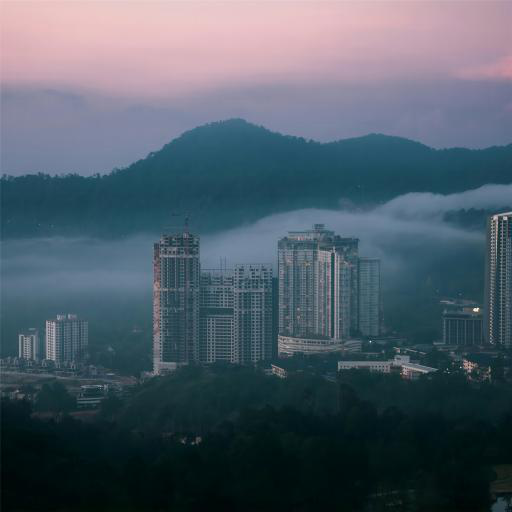}
    \label{fig:61}
  \end{subfigure}
  \begin{subfigure}[b]{0.19\textwidth}
    \centering
    \includegraphics[width=\textwidth]{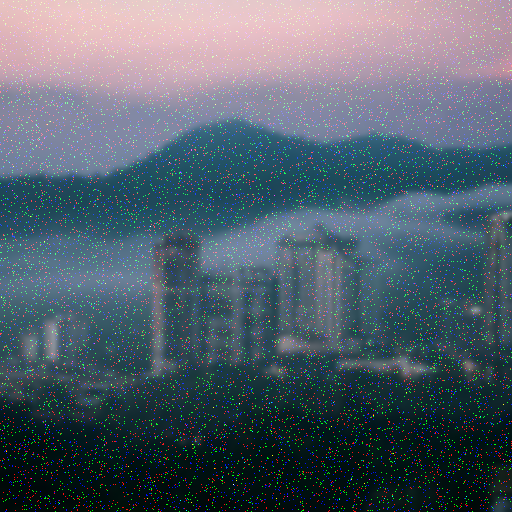}
    \label{fig:62}
  \end{subfigure}
  \begin{subfigure}[b]{0.19\textwidth}
    \centering
    \includegraphics[width=\textwidth]{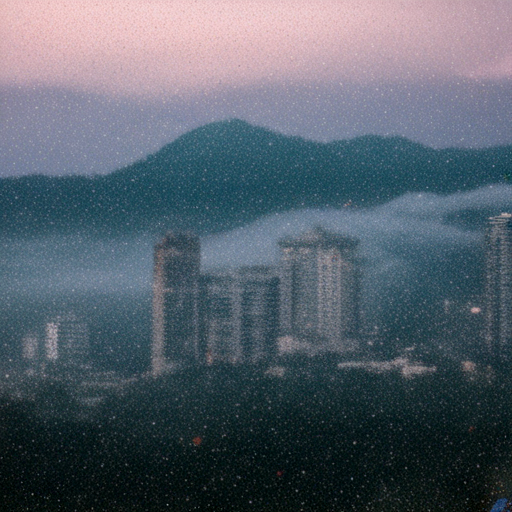}
    \label{fig:63}
  \end{subfigure}
  \begin{subfigure}[b]{0.19\textwidth}
    \centering
    \includegraphics[width=\textwidth]{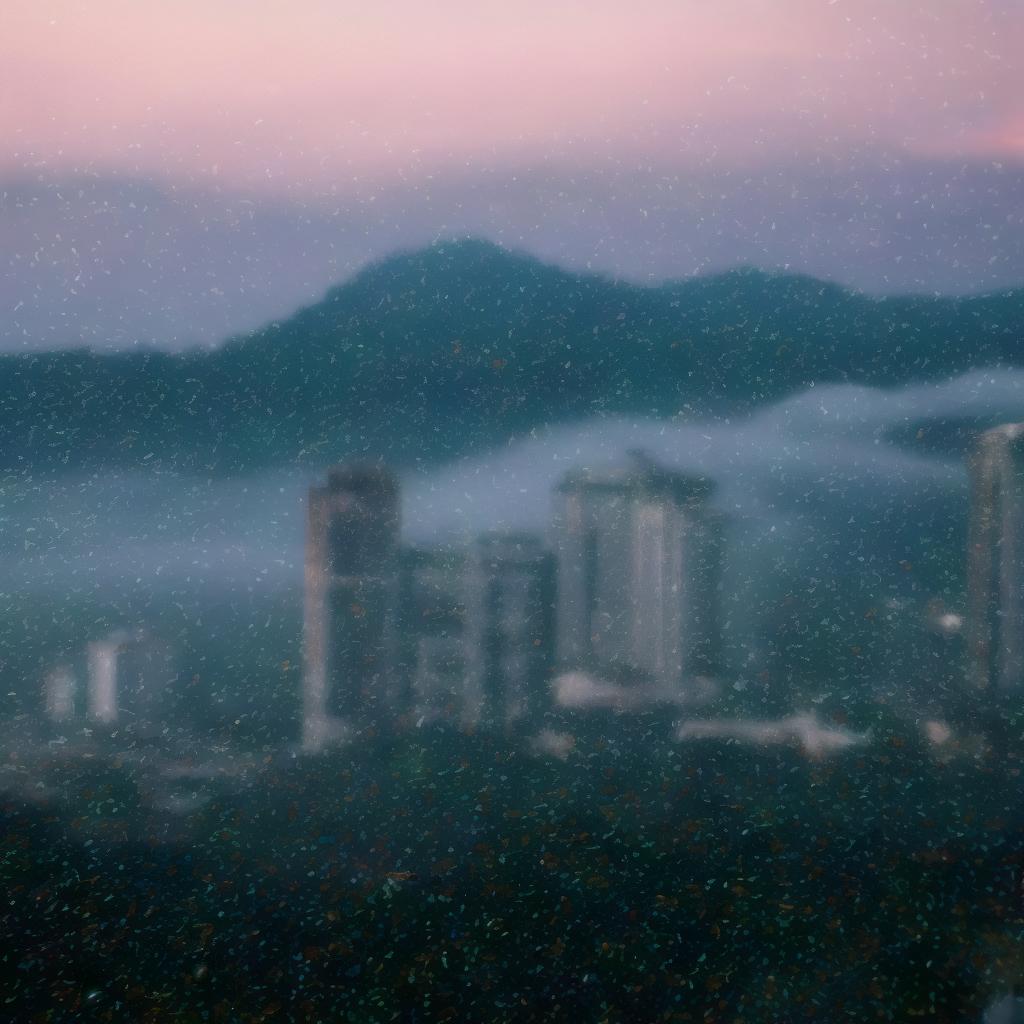}
    \label{fig:64}
  \end{subfigure}
  \begin{subfigure}[b]{0.19\textwidth}
    \centering
    \includegraphics[width=\textwidth]{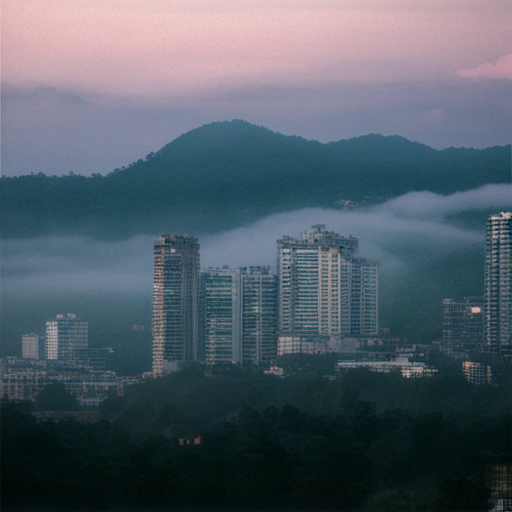}
    \label{fig:65}
  \end{subfigure}

  \vspace{-0.2cm}
  \vspace{-0.2cm}

  \begin{subfigure}[b]{0.19\textwidth}
    \centering
    \includegraphics[width=\textwidth]{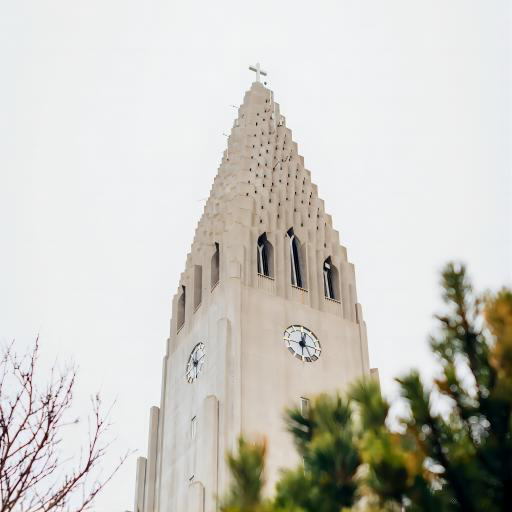}
    \caption{GT}
    \label{fig:71}
  \end{subfigure}
  \begin{subfigure}[b]{0.19\textwidth}
    \centering
    \includegraphics[width=\textwidth]{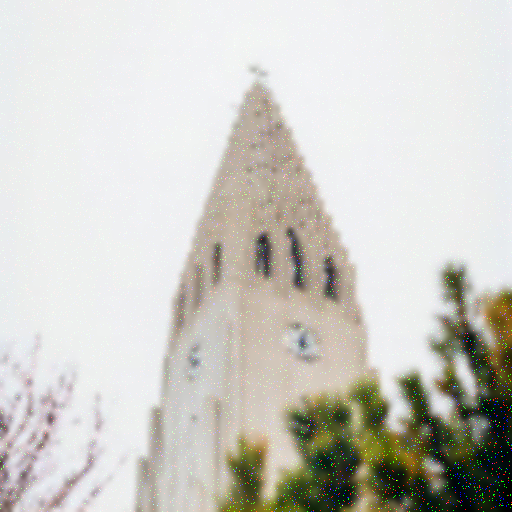}
    \caption{Low-quality image}
    \label{fig:72}
  \end{subfigure}
  \begin{subfigure}[b]{0.19\textwidth}
    \centering
    \includegraphics[width=\textwidth]{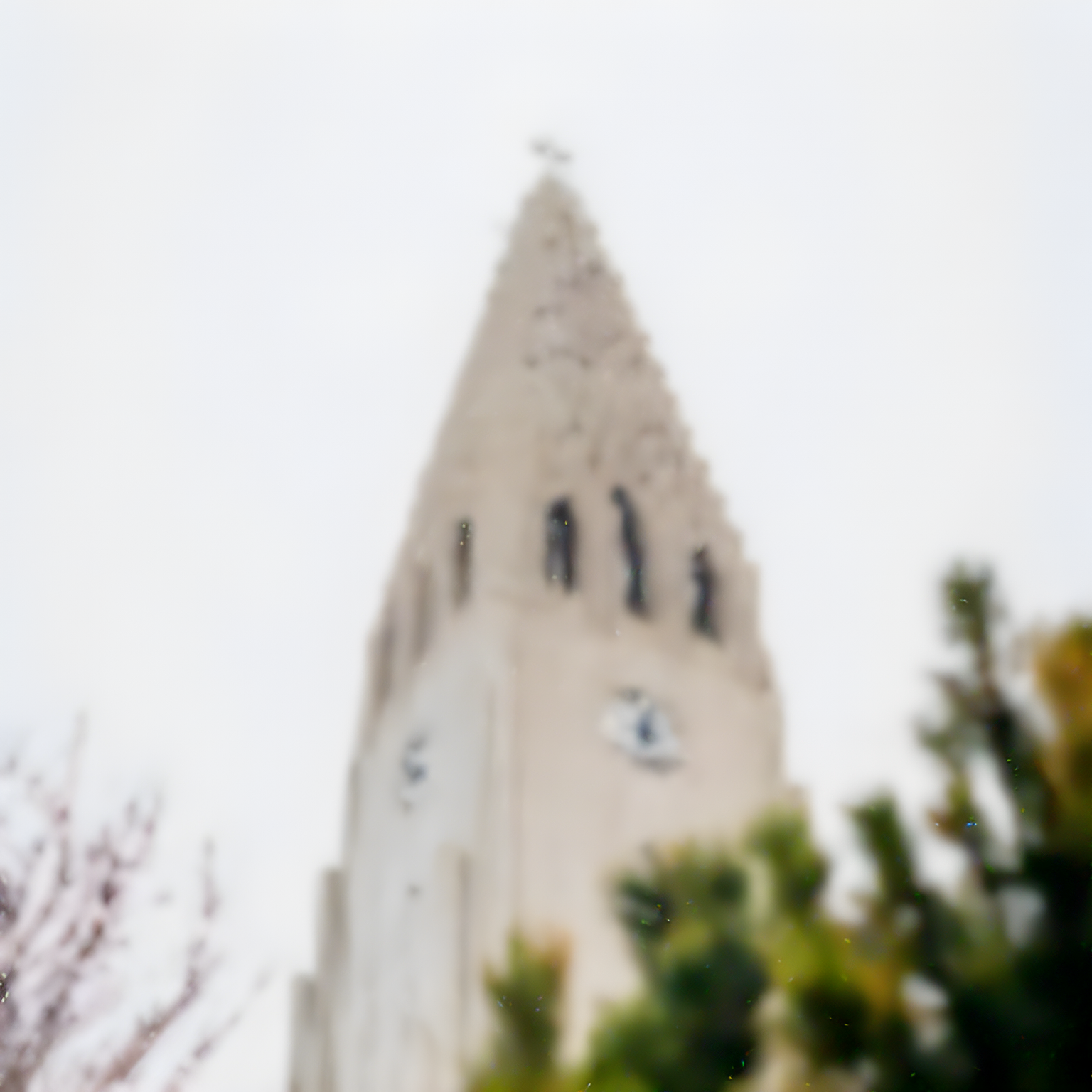}
    \caption{Stable-SR}
    \label{fig:73}
  \end{subfigure}
  \begin{subfigure}[b]{0.19\textwidth}
    \centering
    \includegraphics[width=\textwidth]{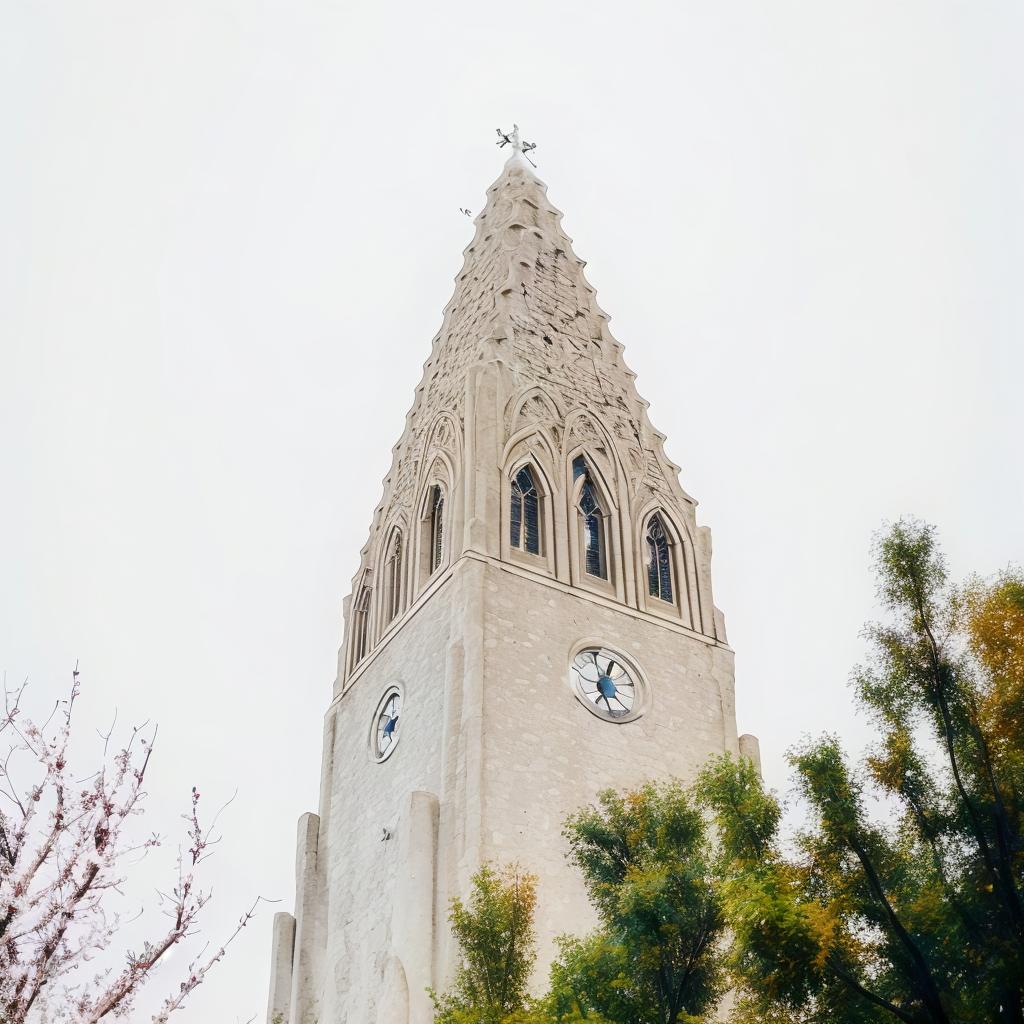}
    \caption{PASD}
    \label{fig:74}
  \end{subfigure}
  \begin{subfigure}[b]{0.19\textwidth}
    \centering
    \includegraphics[width=\textwidth]{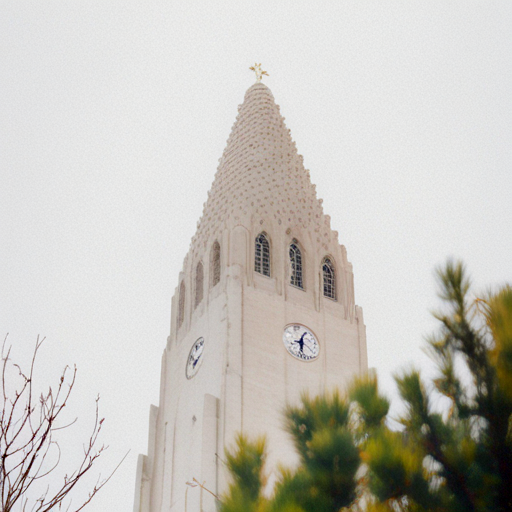}
    \caption{Ours}
    \label{fig:75}
  \end{subfigure}

  \caption{Qualitative comparison with different methods. Our method can accurately restore the texture and details of the corresponding object under challenging degradation. Other methods may have deficiencies in presenting details, such as house, windows and clocks}
  \label{fig:4}
\end{figure*}

\subsection{Comparison with Other Methods}

We also conducted tests on low-quality images and compared them with other models, such as DiffBIR \cite{lin2023diffbir}, Stable-SR \cite{wang2024stablesr}, PASD \cite{yang2023PASD}. We selected the following metrics for quantitative comparison: the  reference metrics PSNR, SSIM, LPIPS.In terms of results, our method achieved the best scores on PSNR and SSIM and LPIPS, indicating that our method has higher perceptual similarity between the restored image and the reference image than other methods.

LoRA reduces the complexity of the model through parameter decomposition, thereby reducing time. As shown in \cref{tab:3}, the comparison between the original method and our method shows that the LoRA method has improved by nearly 7 seconds compared to before. Compared with the other two models, our method still has the shortest time. However, StableSR requires 200 steps to generate a perfect image and consumes a lot of time. This efficiency gain demonstrates the effectiveness of our approach in handling large-scale models. Moreover, the reduction in computational time does not compromise the quality of the generated images, as evidenced by the consistent performance metrics. Moreover, \cref{fig:4}, we can clearly see the differences between the stable SR model and other models, but its performance is not very good. The PASD model performs well in restoring details, such as in case 1. However, PASD has a low ability to restore images with high noise and blur. In case 2, it was unable to restore the windows of distant high-rise buildings and still had noise points in the restored images. In case 3, the restoration of the clock changed its original color.

\section{Conclusion}

In this study, we propose an enhanced image restoration model based on the LoRA module and the SDXL framework. Our method utilizes the advantages of LoRA to fine tune the SDXL model, thereby improving image restoration quality and reducing computation time. Experiments have shown that the proposed model outperforms the original SUPIR model and several other methods in most image degradation scenarios. This indicates that it has better performance and structural fidelity. However, there are still some challenges that need to be addressed. Firstly, when the added blur and noise are too large, the performance of the LoRA enhanced SUPIR model tends to be on par with the original SUPIR model. This indicates that the effectiveness of the LoRA module decreases in extreme degradation scenarios. This observation suggests a potential direction for future research: developing more robust adaptation mechanisms to more effectively handle high-level blurring and noise. In the future, more robust adaptation mechanisms can be developed to more effectively handle high-level blurring and noise. In addition, expanding the image dataset to include a wider variety of real-world images can further enhance the model's ability to handle details.

{
    \small
    \bibliographystyle{ieeenat_fullname}
    \bibliography{main}
}


\end{document}